\documentclass[runningheads]{llncs}
\usepackage{graphicx}
\usepackage{amsmath,amssymb} 
\usepackage{color}
\usepackage[width=122mm,left=12mm,paperwidth=146mm,height=193mm,top=12mm,paperheight=217mm]{geometry}

\usepackage{booktabs}

\newcommand{\eg}{\textit{e.g.}}
\newcommand{\ie}{\textit{i.e.}}
\newcommand{\etal}{\textit{et al.}}
\newcommand{\tc}[1]{\textcolor{red}{#1}}

\usepackage{array}
\newcolumntype{L}[1]{>{\raggedright\let\newline\\\arraybackslash\hspace{0pt}}m{#1}}
\newcolumntype{C}[1]{>{\centering\let\newline\\\arraybackslash\hspace{0pt}}m{#1}}
\newcolumntype{R}[1]{>{\raggedleft\let\newline\\\arraybackslash\hspace{0pt}}m{#1}}
\usepackage[export]{adjustbox}
\usepackage{enumitem}

\begin{document}

\pagestyle{headings}
\mainmatter

\title{Learning Joint Representations of Videos and Sentences with Web Image Search} 

\titlerunning{Learning Joint Representations of Videos and Sentences}

\authorrunning{M. Otani et al.}

\author{Mayu Otani\inst{1} \and Yuta Nakashima\inst{1} \and Esa Rahtu\inst{2} \and \\Janne Heikkil\"a\inst{2} \and Naokazu Yokoya\inst{1}}


\institute{{Graduate School of Information Science, Nara Institute of Science and Technology\\
	\email{ \{otani.mayu.ob9,n-yuta,yokoya\}@is.naist.jp}}\and
    {Center for Machine Vision and Signal Analysis, University of Oulu\\
	\email{ \{erahtu,jth\}@ee.oulu.fi}}
}

\maketitle

\begin{abstract}
Our objective is video retrieval based on natural language queries. In addition, we consider the analogous problem of retrieving sentences or generating descriptions given an input video. Recent work has addressed the problem by embedding visual and textual inputs into a common space where semantic similarities correlate to distances. 
We also adopt the embedding approach, and make the following contributions: First, we utilize web image search in sentence embedding process to disambiguate fine-grained visual concepts. Second, we propose embedding models for sentence, image, and video inputs whose parameters are learned simultaneously. Finally, we show how the proposed model can be applied to description generation.
Overall, we observe a clear improvement over the state-of-the-art methods in the video and sentence retrieval tasks. In description generation, the performance level is comparable to the current state-of-the-art, although our embeddings were trained for the retrieval tasks.

\end{abstract}

\section{Introduction}
During the last decade, the Internet has become an increasingly important distribution channel for videos.
Video hosting services like YouTube, Flickr, and Vimeo have millions of users uploading and watching content every day.
At the same time, powerful search methods have become essential to make good use of such vast databases. By analogy, without textual search tools like Google or Bing, it would be nearly hopeless to find information from the websites. 

Our objective is to study the problem of retrieving video clips from a database using natural language queries. In addition, we consider the analogous problem of retrieving sentences or generating descriptions based on a given video clip. We are particularly interested in learning appropriate representations for both visual and textual inputs. Moreover, we intend to leverage the supporting information provided by the current image search approaches.

This topic has recently received plenty of attention in the community, and papers have presented various approaches to associate visual and textual data. One direction to address this problem is to utilize metadata that can be directly compared with queries. For instance, many web image search engines evaluate the relevance of an image based on the similarity of the query sentence with the user tags or the surrounding HTML text \cite{Datta2008}. For sentence retrieval, Ordonez \etal~\cite{Ordonez2011} proposed to compare an image query and visual metadata with sentences.

While these methods using comparable metadata have demonstrated impressive results, they do not perform well in cases where appropriate metadata is limited or not available. Moreover, they rely strongly on the assumption that the associated visual and textual data in the database is relevant to each other. These problems are more apparent in the video retrieval task since video distribution portals like YouTube often provide less textual descriptions compared to other web pages. Furthermore, available descriptions (e.g. title) often cover only a small portion of the entire visual content in a video.  

\begin{figure}[t!]
\centering
\includegraphics[clip, width=0.9\linewidth]{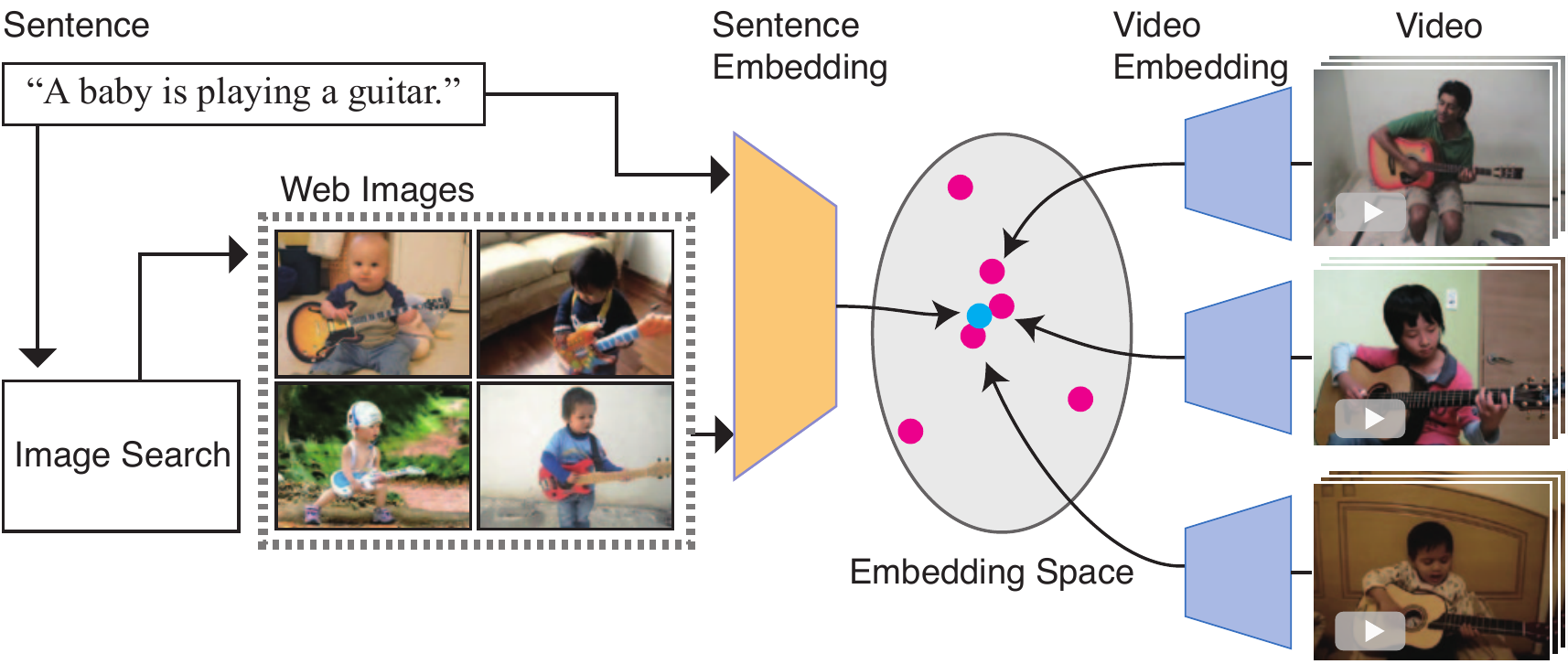}
\caption{An overview of our approach. Left side illustrates the image search results for a query ``A baby is playing a guitar''. Images highlight evidence of objects (``baby'', ``guitar'') and actions (``playing''). Right side shows the most relevant videos in the YouTube dataset \cite{Chen2011} obtained by ranking the clips according to Euclidean distance to the query sentence in an embedding space.  \label{fig:overview} }
\end{figure}

An alternative approach would be to compare textual and visual inputs directly. In many approaches, this is enabled by embedding the corresponding representations into a common vector space in such a way that the semantic similarity of the original inputs would be directly reflected in their distance in the embedding space (Fig. \ref{fig:overview}). Recent work \cite{Socher2013,Kiros2015} has proposed deep neural network models for performing such embeddings. The results are promising, but developing powerful joint representations still remains a challenge. 

In this paper, we propose a new embedding approach for sentence and video inputs that combines the advantages of the metadata-based web image search and deep neural network-based representation learning. More precisely, we use a standard search engine to obtain a set of supplementary images for each query sentence. Then, we pass the sentence and the retrieved images to a two-branch neural network that produces the sentence embedding. The video inputs are embedded into the same space using another neural network. The network parameters are trained jointly so that videos and sentences with similar semantic content are mapped to close points. Figure \ref{fig:overview} illustrates the overall architecture of our approach. The experiments indicate a clear improvement over the current state-of-the-art baseline methods.

Our main contributions are as follows:
\begin{itemize}
\item We present an embedding approach for video retrieval that incorporates web image search results to disambiguate fine-grained visual concepts in query sentences.
\item We introduce neural network-based embedding models for video, sentence, and image inputs whose parameters can be learned jointly. Unlike previous work that uses only videos and sentences, we utilize a sentence and corresponding web images to compute the sentence embedding.
\item We demonstrate a clear improvement over the state-of-the-art in the video and sentence retrieval tasks with the YouTube dataset \cite{Chen2011}.
\item We demonstrate description generation as an example of possible applications of our video embeddings. We observed that the performance is comparable with the state-of-the-art. This indicates that video contents are efficiently encoded into our video embeddings.
\end{itemize}

\section{Related Work}
\subsubsection*{Visual and Language Retrieval:}
Due to the explosive growth of images and videos on the web, visual retrieval has become a hot topic in computer vision and machine learning \cite{Datta2008,Maybank2011}.
Several recent approaches for joint representation leaning enable direct comparison among different multimodalities.
Farhadi \etal~\cite{Farhadi2010} introduced triplets of  labels on object, action, and scene as joint representations for images and sentences.
Socher \etal~\cite{Socher2013} proposed to embed representations of images and labels into a common embedding space.
For videos, the approach proposed by Lin \etal~\cite{Lin2014} associates a parsed semantic graph of a query sentence and visual cues based on object detection and tracking.

The recent success of deep convolutional neural networks (CNNs) together with large-scale visual datasets \cite{Rashtchian2010,capeval2015,Russakovsky2015} has resulted in several powerful representation models for images \cite{Donahue2014,Wang2015,Yao2015}.
These CNN-based methods have been successfully applied to various types of computer vision tasks, such as object detection \cite{Girshick2014,NIPS2015_5638}, video summarization \cite{Gygli2015}, and image description generation \cite{Vinyals2015,Fang2015}.

Deep neural networks have also been used in the field of natural language processing \cite{Le2014,Kiros2015}. For example, Kiros \etal~\cite{Kiros2015} proposed sentence representation learning based on recurrent neural networks (RNNs). 
They also demonstrated image and sentence retrieval by matching sentence and image representations with jointly leaned linear transformations.

Representation learning using deep neural networks is explored in many tasks \cite{Chopra2005,Lin2015,Frome2013,Karpathy2014,Xu2015,Zhu2015}.
Frome \etal~\cite{Frome2013} proposed image classification by computing similarity between joint representations of images and labels, and Zhu \etal~\cite{Zhu2015} addressed alignment of a movie and sentences in a book using joint representations for video clips and sentences.
Their approach also computes similarity between sentences and subtitles of video clips to improve the alignment of video clips and sentences.

Our approach is the closest to work by Xu \etal~\cite{Xu2015}. They represent a sentence by a subject, verb, and object (SVO) triplet, and embed sentences as well as videos to a common vector space using deep neural networks.
The main difference between ours and the work \cite{Xu2015} is the use of an RNN to encode a sentence and supplementary web images.
The use of an RNN enables our model to encode all words in a sentence and capture details of the sentence, such as an object's attributes and scenes, together with corresponding web images.

\subsubsection*{Exploiting Image Search:}
The idea of exploiting web image search is adopted in many tasks, including object classification \cite{Fergus2005} and video summarization \cite{Song2015}.
These approaches collect a vast amount of images from the web and utilize them to extract canonical visual concepts.
Recent label prediction for images by Johnson \etal~\cite{Johnson2015} infers tags of target images by mining relevant Flickr images based on their metadata, such as user tags and photo groups curated by users.
The relevant images serve as priors on tags for the target image.
A similar motivation drives us to utilize web images for each sentence, which can disambiguate visual concepts of the sentence and highlight relevant target videos.

\section{Proposed Approach}
\begin{figure}[t!]
\centering
\includegraphics[clip, width=0.9\linewidth]{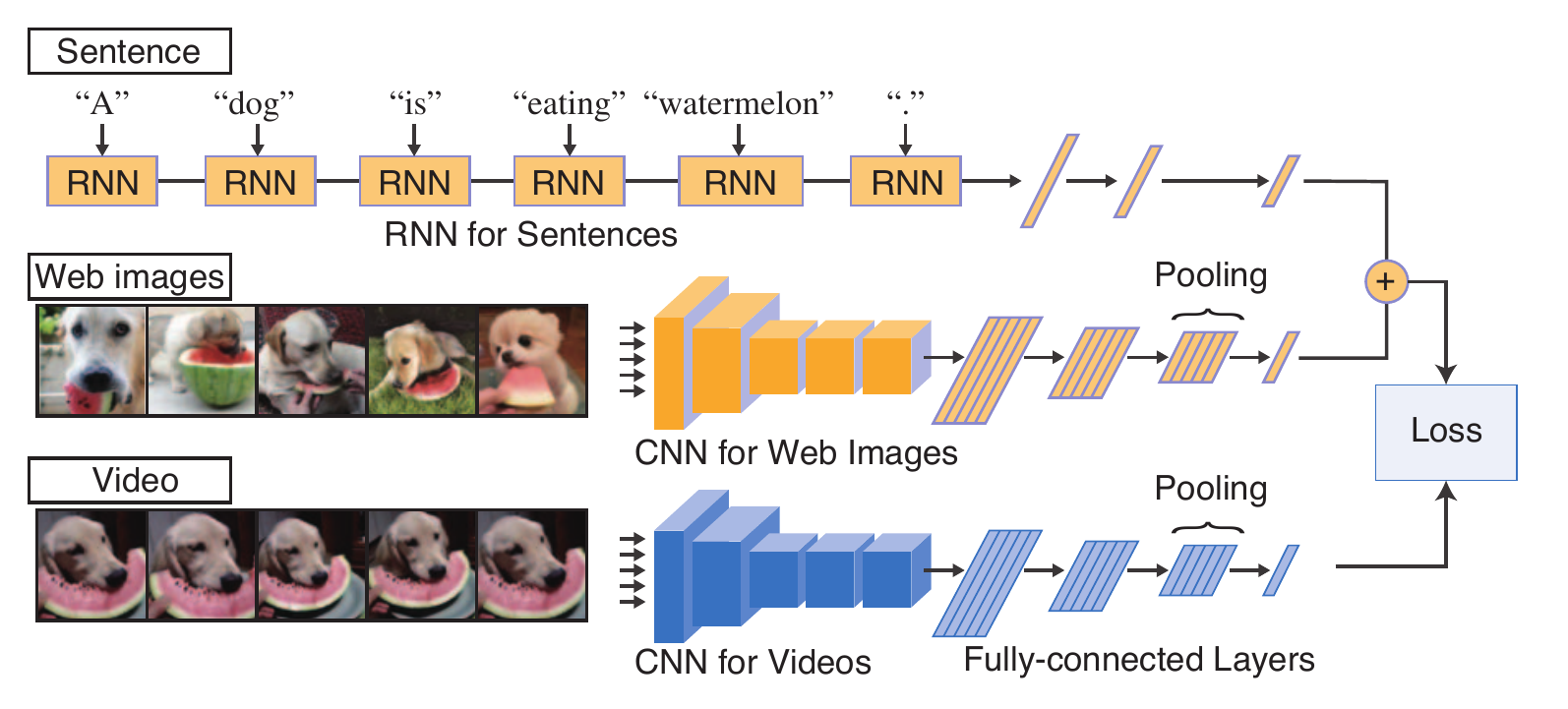}
\caption{Illustration of our video and sentence embedding. The orange component is the sentence embedding model that takes a sentence and corresponding web images as input. Video embedding model is denoted by the blue component. \label{fig: input_model}}
\end{figure}
We propose neural network-based embedding models for the video and sentence retrieval tasks.
In order to enhance the sentence embedding, we retrieve relevant web images that are assumed to disambiguate semantics of the sentence.
For example, the word ``keyboard'' can be interpreted as a musical instrument or an input device for computers.
If the word comes with ``play,'' the meaning of ``keyboard'' narrows down to a musical instrument.
This means that a specific combination of words can reduce the possible visual concepts relevant to the sentence, which may not be fully encoded even with the state-of-the-art RNN-based approach like \cite{Kiros2015}.

We propose to take this into account by using web image search results.
Since most image search engines use surrounding text to retrieve images, we can expect that they are responsive to such word combinations.
Consequently, we retrieve web images using the input sentence as a query and download the results.
The web images are fused with the input sentence by applying a two-branch neural network as shown in Fig.~\ref{fig: input_model}.
Videos are also encoded by applying a neural network-based video embedding model.
Relevance between sentence and video inputs is directly computed in the common embedding space using Euclidean distances.
We jointly train our embedding models using video-sentence pairs by minimizing the contrastive loss \cite{Chopra2005}.

\subsection{Video Embedding}
We extract frames from a video at 1 fps as in \cite{Xu2015} and feed them to a CNN-based video embedding model.
In our approach, we employ two CNN architectures: 19-layer VGG \cite{Simonyan2015} and GoogLeNet \cite{Szegedy2015}, both of which are pre-trained on ImageNet \cite{Russakovsky2015}.
We replace the classifier layer in each model with two fully-connected layers.
Specifically, we compute activations of the VGG's fc7 layer or the GoogLeNet's inception 5b layer and feed them to additional embedding layers.

Let $X = \{x_i \mid  i = 1, \ldots, M\}$ be a set of frames $x_i$, and $\mathrm{CNN}(x_i) \in \mathrm{R}^{d_{\mathrm{v}}}$ be an activation of a CNN ($d_{\mathrm{v}}$=4,096 for VGG, and $d_{\mathrm{v}}$=1,024 for GoogLeNet).
The video embedding  $\phi_{\mathrm{v}}(X) \in \mathrm{R}^{d_\mathrm{e}}$ is computed by:
\begin{equation}
\phi_{\mathrm{v}}(X) = \frac{1}{M} \sum_{x_i \in X}\ \tanh(W_{\mathrm{v}_2} \tanh (W_{\mathrm{v}_1} \mathrm{CNN}(x_i) + b_{\mathrm{v}_1})+b_{\mathrm{v}_2}).
\end{equation}
 Here, $W_{\mathrm{v}_1} \in \mathrm{R}^{d_\mathrm{h}\times d_{\mathrm{v}}}$, $b_{\mathrm{v}_1} \in \mathrm{R}^{d_\mathrm{h}}$, $W_{\mathrm{v}_2} \in \mathrm{R}^{d_\mathrm{e} \times d_\mathrm{h}}$, and  $b_{\mathrm{v}_2} \in \mathrm{R}^{d_\mathrm{e}}$ are the learnable parameters of the fully-connected layers.
 
 \subsection{Sentence and Web Image Embedding}
The sentence embedding model consists of two branches that merge the outputs of a CNN-based network for web images and an RNN-based network for a sentence.
Before computing the sentence embedding, we download top-$K$ results of web image search with the input sentence as a query.
Let  $Z = \{z_j \mid j = 1, \ldots, K\}$ be a set of web images.
We utilize the same architecture as the video embedding and compute an intermediate representation $e_{\mathrm{z}}\in \mathrm{R}^{d_\mathrm{e}}$ that integrates the web images as:
\begin{equation}
e_\mathrm{z}  = \frac{1}{K}\sum_{z_j\in Z} \tanh(W_{\mathrm{z}_2} \tanh (W_{\mathrm{z}_1} \mathrm{CNN}(z_j) + b_{\mathrm{z}_1})+b_{\mathrm{z}_2}),
\end{equation}
where $W_{\mathrm{z}_1} \in \mathrm{R}^{d_\mathrm{h}\times d_{\mathrm{v}}}$, $b_{\mathrm{z}_1} \in \mathrm{R}^{d_\mathrm{h}}$, $W_{\mathrm{z}_2} \in \mathrm{R}^{d_\mathrm{e} \times d_\mathrm{h}}$, and $b_{\mathrm{z}_2} \in \mathrm{R}^{d_\mathrm{e}}$ are the leanable parameters of the two fully-connected layers.

We encode sentences into vector representations using skip-thought that is an RNN pre-trained with a large-scale book corpus \cite{Kiros2015}.
Let $Y=\{y_t \mid t=1,\ldots, T_Y\}$ be the input sentence, where $y_t$ is the $t$-th word in the sentence, and $T_Y$ is the number of words in the sentence $Y$.
Skip-thought takes a sequence of word vectors $w_t \in \mathrm{R}^{d_\mathrm{w}}$ computed from a word input $y_t$ as in \cite{Kiros2015} and produces hidden state $h_t \in \mathrm{R}^{d_{\mathrm{s}}}$ at each time step $t$ as:
\begin{eqnarray}
r_t &=& \sigma (W_\mathrm{r} w_{t}+ U_\mathrm{r} w_{t-1}),\\
i_t &=& \sigma (W_\mathrm{i} w_{t}+ U_\mathrm{i} h_{t-1}),\\
a_t &=& \tanh (W_\mathrm{a} w_{t} + U_\mathrm{a} (r_t \odot h_{t-1})), \\
h_t &=& (1-i_t) \odot h_{t-1} + i_t \odot a_t,
\end{eqnarray}
where $\sigma$ is the sigmoid activation function, and $\odot$ is the component-wise product.
The parameters $W_\mathrm{r}, W_\mathrm{i}, W_\mathrm{a},U_\mathrm{r}, U_\mathrm{i}$, and $U_\mathrm{a}$ are $d_{\mathrm{s}} \times d_{\mathrm{w}}$ matrices.
Sentence $Y$ is encoded into the hidden state after processing the last word $w_{T_Y}$, \ie, $h_{T_Y}$.
We use combine-skip in \cite{Kiros2015}, which is a concatenation of outputs from two separate RNNs trained with different datasets.
We denote the output of combine-skip from sentence $Y$ by $s_{Y} \in \mathrm{R}^{d_\mathrm{c}}$, where $d_\mathrm{c}$=4,800.

We also compute an intermediate representation $e_\mathrm{s}$ for sentence $Y$ as:
\begin{equation}
e_\mathrm{s} =  \tanh(W_{\mathrm{s}_2} \tanh (W_{\mathrm{s}_1} s_{Y} + b_{\mathrm{s}_1})+b_{\mathrm{s}_2}),
\end{equation}
where $W_{\mathrm{s}_1} \in \mathrm{R}^{d_\mathrm{h}\times d_{\mathrm{c}}}$, $b_{\mathrm{s}_1} \in \mathrm{R}^{d_\mathrm{h}}$, $W_{\mathrm{s}_2} \in \mathrm{R}^{d_\mathrm{e}\times d_\mathrm{h}}$, and $b_{\mathrm{s}_2} \in \mathrm{R}^{d_\mathrm{e}}$ are the learnable parameters of sentence embedding.

Once the outputs $e_\mathrm{s}$ and $e_\mathrm{z}$ of each branch in our sentence embedding model are computed, they are merged into a sentence embedding $\phi_\mathrm{s}(Y,Z)$ as:
\begin{equation}
\phi_{\mathrm{s}}(Y, Z) = \frac{1}{2} (e_\mathrm{s}+e_\mathrm{z}).
\end{equation}
By this simple mixture of $e_\mathrm{s}$ and $e_\mathrm{z}$, the sentence and web images directly influence the sentence embedding.

\subsection{Joint Learning of Embedding Models}
We jointly train both embedding  $\phi_{\mathrm{v}}$ and $\phi_{\mathrm{s}}$ using pairs of videos and associated sentences in a training set by minimizing the contrastive loss function \cite{Chopra2005}.
In our approach, the contrastive loss decreases when embeddings of videos and sentences with similar semantics get closer to each other in the embedding space, and those with dissimilar semantics get farther apart.

The training process requires a set of positive and negative video-sentence pairs.
A positive pair contains a video and a sentence that are semantically relevant, and a negative pair contains irrelevant ones.
Let $\{(X_n, Y_n)\mid n=1, \ldots, N\}$ be the set of positive pairs.
Given a positive pair $(X_n, Y_n)$, we sample irrelevant sentences $\mathcal{Y}'_n = \{Y'_f \mid f = 1, \ldots, N_c\}$ and videos $\mathcal{X}'_n = \{X'_g \mid g = 1, \ldots, N_c\}$ from the training set, which are used to build two sets of negative pairs $\{(X_n, Y'_f)\mid Y'_f \in \mathcal{Y}'_n\}$ and $\{(X'_g, Y_n)\mid X'_g \in \mathcal{X}'_n\}$.
In our approach, we set the size of negative pairs $N_c$ to 50.
We train the parameters of embedding  $\phi_{\mathrm{v}}$ and $\phi_{\mathrm{s}}$ by minimizing the contrastive loss defined as:
\begin{eqnarray}
Loss(X_n, Y_n) =&\frac{1}{1+2N_c}&\biggl\{ d(X_n, Y_n)\nonumber \\ 
&+&\sum_{Y'_f \in \mathcal{Y}'_n} \mathrm{max}(0, \alpha - d(X_n, Y'_f))\nonumber \\ 
&+&\sum_{X'_g \in \mathcal{X}'_n} \mathrm{max}(0, \alpha -d(X'_g, Y_n) ) \biggr\},\\
d(X_i, Y_j) &=& ||\phi_{\mathrm{v}}(X_i) - \phi_{\mathrm{s}}(Y_j, Z_j)||_2^2,
\end{eqnarray}
where $Z_n$ is the web images corresponding to sentence $Y_n$.
The hyperparameter $\alpha$ is a margin.
Negative pairs with smaller distances than $\alpha$ are penalized.
Margin $\alpha$ is set to the largest distance of positive pairs before training so that most negative pairs influence the model parameters at the beginning of training.
\begin{figure}[t!]
\includegraphics[clip, width=\linewidth]{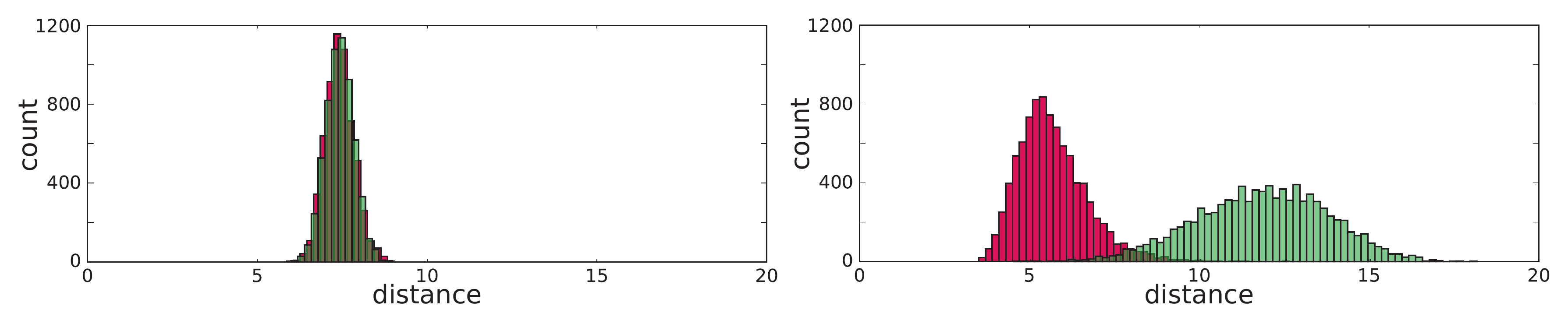}\\
\includegraphics[clip, width=\linewidth]{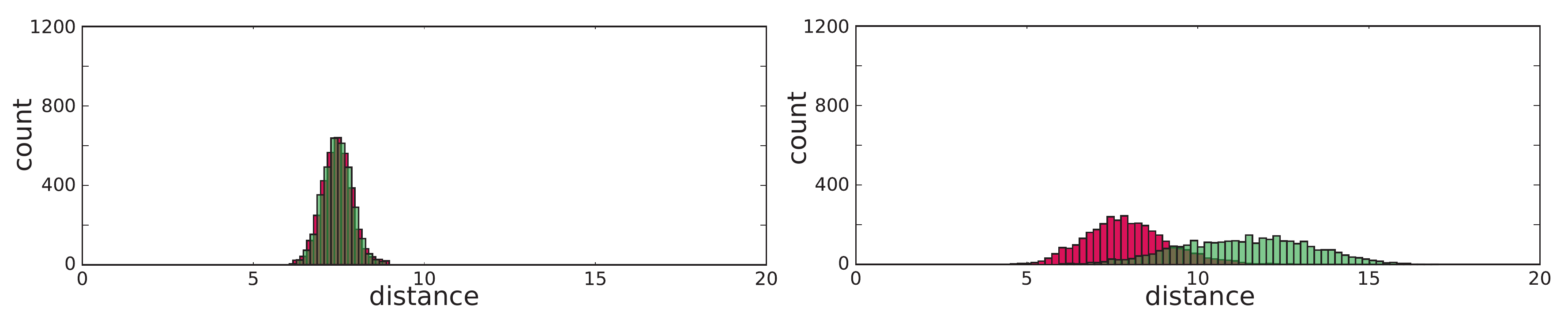}
\caption{Histograms of pairwise distances before training (left) and after training (right). Top row: Histograms of the training set. Bottom row: Histograms of the test set. Red represents positive pairs, and green represents negative pairs. \label{fig:dist_hist}}
\end{figure}

\begin{figure}[t!]
\centering
\includegraphics[clip, width=0.9\linewidth]{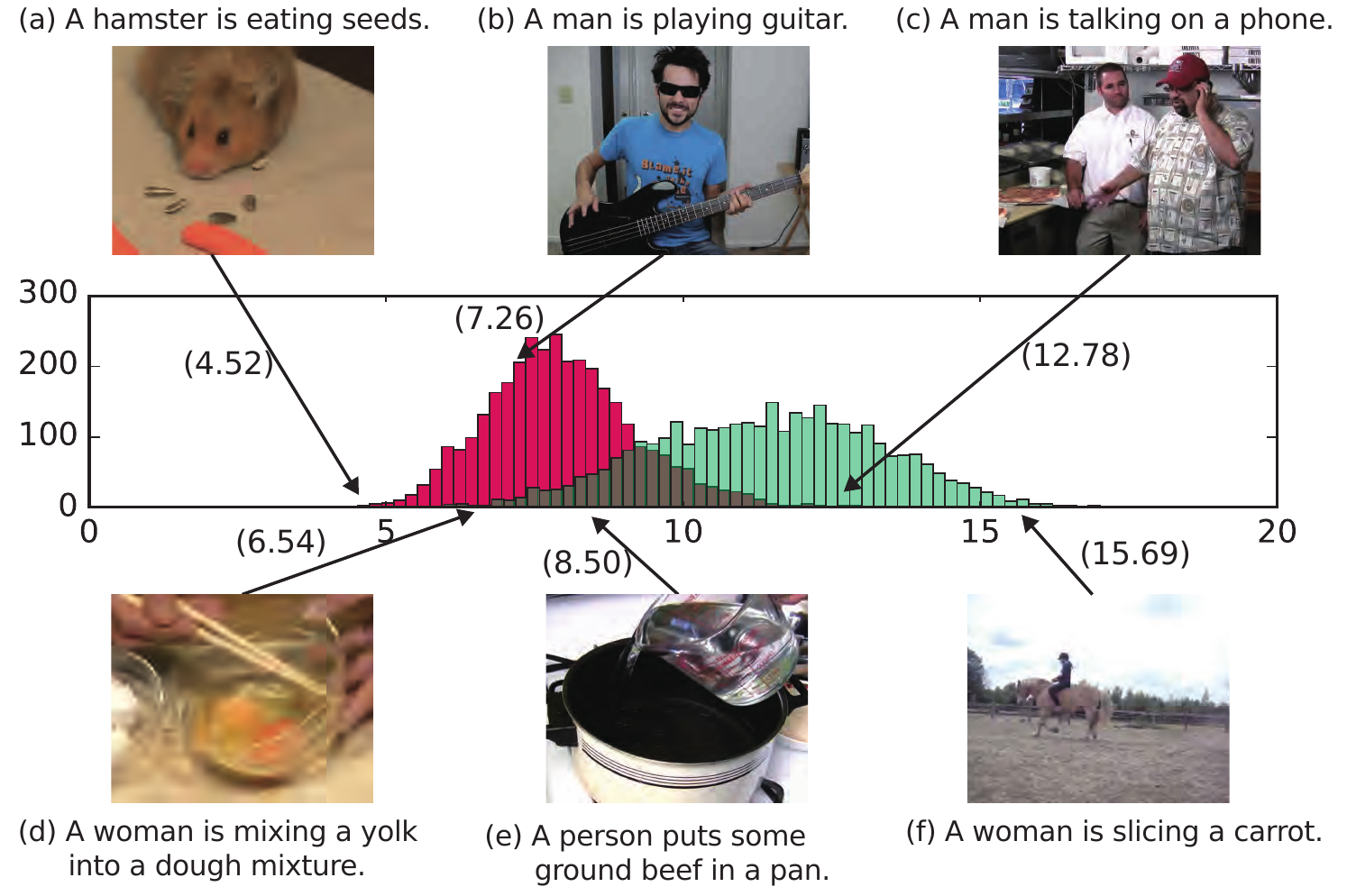}
\caption{Examples of positive (a)--(c) and negative (d)--(f) pairs in the test set with corresponding distances. The values ($\cdot$) are distances of the pairs. The plot is the histograms of distances of positive (red) and negative (green) pairs.  \label{fig:sample_pair}}
\end{figure}

Figure \ref{fig:dist_hist} shows the histograms of distances of positive and negative pairs before and after training.
The initial distance distributions of positive and negative pairs overlap.
After training, the distributions are pulled apart.
This indicates that the training process encourages videos and sentences in positive pairs to be mapped to closer points and those in negative ones to farther points.

The examples of positive and negative pairs in our test set with corresponding distances are shown in Fig. \ref{fig:sample_pair}.
The positive pair (a) and (b) are easy cases, in which sentences explicitly describe the video contents.
The pair (c) is an example of hard cases.
The sentence includes ``a man'' and ``phone'', but the video actually shows two men, and a phone is occluded by a hand.

The pairs (d) and (e) are hard negative cases.
The pair (d) shows partial matches of contents, such as the action ``mixing'' and the object ``yolk.''
Another negative pair (e) has a video and a sentence about cooking, although there is disagreement about details.
As shown in these examples, the closer a video and a sentence are located in the embedding space, the more relevant they are. 
More examples can be found in the supplementary material.

\section{Retrieval Experiments}
\subsection{Implementation Detail}
With 19-layer VGG, the hidden layer size $d_\mathrm{h}$ of embedding  $\phi_{\mathrm{v}}$ and $\phi_{\mathrm{s}}$ was set to 1,000 and the dimension of the embedding space $d_\mathrm{e}$ was set to 300.
For model using GoogLeNet, we used $d_\mathrm{h}=600$ and  $d_\mathrm{e}=300$.

We implemented our model using Chainer \cite{chainer_learningsys2015}.
We used Adam \cite{Kingma2015} for optimization with a learning rate of $1 \times 10^{-4}$.
The parameters of the CNNs and skip-thought were fixed.
We applied dropout with a ratio of 0.5 to the input of the first and second layers of $\phi_{\mathrm{v}}$ and $\phi_{\mathrm{s}}$.
Our models were trained for 15 epochs, and their parameters were saved at every 100 updates.
We took the model parameters whose performance was the best on the validation set.

\subsection{Experimental Setup}
\subsubsection*{Dataset:}
We used the YouTube dataset \cite{Chen2011} consisting of 80K English descriptions for 1,970 videos.
We first divided the dataset into 1,200, 100, and 670 videos for training, validation, and test, respectively, as in \cite{Yao2015,Xu2015,Guadarrama2013}.
Then, we extracted five-second clips from each original video in a sliding-window manner.
As a result, we obtained 8,001, 628, and 4,499 clips for the training, validation, and test sets, respectively.
For each clip, we picked five ground truth descriptions out of those associated with its original video.

We collected top-5 image search results for each sentence using the Bing image search engine.
We used a sentence modified by lowercasing and punctuation removal as a query.
In order to eliminate cartoons and clip art, the image type was limited to photos using Bing API.

\subsubsection{Video Retrieval:}
Given a video and a query sentence, we extracted five-second video clips from the video and computed Euclidean distances from the query to the clips.
We used their median as the distance of the original video and the query.
We ranked the videos based on the distance to each query and recorded the rank of the ground truth video.
Since the test set has 670 videos, the probability of bringing the ground truth video at top-1 by random ranking is about 0.14\%.

\subsubsection{Sentence Retrieval:}
For the sentence retrieval task, we ranked sentences for each query video.
We computed the distances between a sentence and a query video in the same way as the video retrieval task.
Note that each video has five ground truth sentences; thus, we recorded the highest rank among them.
The test set has 3,500 sentences.

\subsubsection{Evaluation Metrics:}
We report recall rates at top-1, -5, and -10, the average and median rank, which are standard metrics employed in the retrieval evaluation.
We found that some videos in the dataset had sentences whose semantics were almost the same (\eg, ``A group of women is dancing'' and ``Women are dancing''). 
For the video that is annotated with one of such sentences, the other sentence is treated as incorrect with the recall rates, which does not agree with human judges. 
Therefore, we employed additional evaluation metrics widely used in the description generation task, \ie, CIDEr, BLUE@4, and METEOR \cite{capeval2015}.
They compute agreement scores in different ways using a retrieved sentence and a set of ground truth ones associated with a query video.
Thus, these metrics give high scores for semantically relevant sentences even if they are not annotated to a query video.
We computed the scores of the top ranked sentence for each video using the evaluation script provided in the Microsoft COCO Evaluation Server \cite{capeval2015}.
In our experiments, all ground truth descriptions for each original video are used to compute these scores.

\subsection{Effects of Each Component of Our Approach}
\begin{table}[t!]
\small
\centering
\caption{Video and sentence retrieval results. R@$K$ is recall at top $K$ results (higher values are better). aR and mR are the average and median of rank (lower values are better). Bold values denotes best scores of each metric.}
\label{tbl:ranking_score1}
\begin{tabular}{@{}l|ccccc|ccccc@{}}
\hline
\multicolumn{1}{c|}{}                & \multicolumn{5}{c|}{Video retrieval}                                              & \multicolumn{5}{c}{Sentence retrieval}                                            \\ \hline
\multicolumn{1}{c|}{\textbf{Models}} & \textbf{R@1}  & \textbf{R@5}   & \textbf{R@10}  & \textbf{aR} & \textbf{mR}  & \textbf{R@1}  & \textbf{R@5}   & \textbf{R@10}  & \textbf{aR} & \textbf{mR}  \\ \hline
Random Ranking                       & 0.14          & 0.79           & 1.48           & 335.92         & 333         & 0.22          & 0.69           & 1.32           & 561.32         & 439         \\ \hline
VGG+VS                               & 6.12          & 21.88          & 33.22          & 58.98          & 24          & 7.01          & 18.66          & 27.16          & 131.33         & 35          \\
VGG+VI                               & 4.03          & 13.70          & 21.40          & 94.62          & 48          & 5.67          & 17.91          & 28.21          & 116.86         & 38          \\
VGG+$\mathrm{ALL_{1}}$                  & 6.48          & 20.15          & 30.51          & 59.53          & 26          & \textbf{10.60}         & 25.22          & 36.42          & 85.90          & 21          \\
VGG+$\mathrm{ALL_{2}}$                 & 5.97          & 21.31          & 32.54          & 56.01          & 24          & 8.66          & 22.84          & 33.13          & 100.14         & 29          \\
GoogLeNet+VS                         & 7.49          & 22.84          & 33.10          & 54.14          & 22          & 8.51          & 21.34          & 30.45          & 114.66         & 33          \\
GoogLeNet+VI                         & 4.24          & 16.42          & 24.96          & 84.48          & 41          & 6.87          & 17.31          & 30.00          & 96.78          & 30          \\
GoogLeNet+ALL$_1$            & 5.52          & 18.93          & 28.90          & 60.38          & 28          & 9.85 & \textbf{27.01} & \textbf{38.36} & \textbf{75.23} & \textbf{19} \\
GoogLeNet+ALL$_2$            & \textbf{7.67} & \textbf{23.40} & \textbf{34.99} & \textbf{49.08} & \textbf{21} & 9.85 & 24.18          & 33.73          & 85.16          & 22          \\
\hline
ST \cite{Kiros2015}                         & 2.63          & 11.55          & 19.34          & 106.00 &    51     & 2.99          & 10.90          & 17.46          & 241.00     &   77  \\
DVCT \cite{Xu2015}                                 &         -     &       -        &        -       & 224.10          & -&        -      &           -    &         -      & 236.27   &     -  \\
\hline
\end{tabular}
\end{table}

\begin{table}[t!]
\small
\centering
\caption{Evaluated scores of retrieved sentences. All values are reported in percentage (\%). Higher scores are better.}
\label{tbl:agreement_score1}
\begin{tabular}{@{}l|C{1.8cm}C{1.8cm}C{1.8cm}@{}}
\hline
\multicolumn{1}{c|}{\textbf{Models}} & \textbf{CIDEr} & \textbf{BLEU}  & \textbf{METEOR} \\ \hline
VGG+VS       & 30.44          & 27.16          & 25.74           \\
VGG+VI     & 29.00          & 22.42          & 22.99           \\
VGG+ALL$_1$          & 42.52          & \textbf{30.81}          & \textbf{27.77}  \\
VGG+ALL$_2$           & 32.56          & 27.39          & 26.58           \\
GoogLeNet+VS & 33.82          & 26.97          & 25.99           \\
GoogLeNet+VI & 35.08          & 24.56          & 24.16           \\
GoogLeNet+ALL$_1$     & \textbf{43.52} & 29.99 & 27.48           \\
GoogLeNet+ALL$_2$     & 38.08          & 29.28          & 26.50           \\
\hline
\end{tabular}
\end{table}

In order to investigate the influence of each component of our approach, we tested some variations of our full model.
The scores of the models on the video and sentence retrieval tasks are shown in Table \ref{tbl:ranking_score1}.
Our full model is denoted by ALL$_2$.
ALL$_1$ is a variation of ALL$_2$ that computes embeddings with one fully-connected layer with the unit size of $d_\mathrm{e}$.
Comparison between ALL$_1$ and ALL$_2$ indicates that the number of fully-connected layers in embedding is not essential.

In order to evaluate the contributions of web images, we trained a model that does not use web images, \ie, an embedding of a sentence $Y$ is computed by $\phi_{\mathrm{s}}(Y) = e_{\mathrm{s}}$.
We denote this model by VS.
VGG+ALL$_2$ had better average rank than VGG+VS, and comparison between GoogLeNet+ALL$_2$ and GoogLeNet+VS also shows a clear advantage of incorporating web images.

We also tested a model without sentences, which is denoted by VI.
It computes an embedding of web images by $\phi_{\mathrm{s}}(Z) = e_{\mathrm{z}}$.
We investigated the effect of using both sentences and web images by comparing VI to our full model ALL$_2$.
The results show that sentences are necessary.
The comparison between VI and VS also indicates that sentences provide main cues for the retrieval task.

The scores of retrieved sentences computed by CIDEr, BLEU@4, and METEOR are shown in Table \ref{tbl:agreement_score1}.
In all metrics, our model using both sentences and web images (ALL$_1$ and ALL$_2$) outperformed to other models (VS and VI).
In summary, contributions by sentences and web images were non-trivial, and the best performance was achieved by using both of them.

\begin{figure}[t!]
\centering
\begin{tabular}{p{.45\linewidth}|p{.27\linewidth}|p{.27\linewidth}}
\multicolumn{1}{c}{\textbf{Query}} & \multicolumn{1}{c}{\textbf{GoogLeNet+VS}} & \multicolumn{1}{c}{\textbf{GoogLeNet+ALL$_2$}} \\

(1) A man is playing a keyboard.&&\\
\includegraphics[width=1cm]{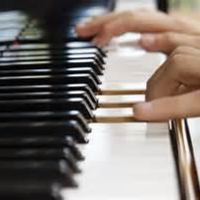}
\includegraphics[width=1cm]{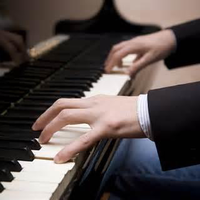}
\includegraphics[width=1cm]{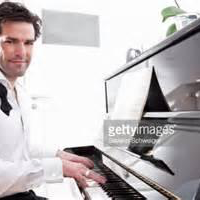}
\includegraphics[width=1cm]{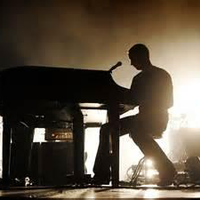}
\includegraphics[width=1cm]{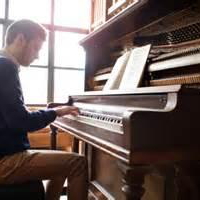}
&
\includegraphics[width=1cm]{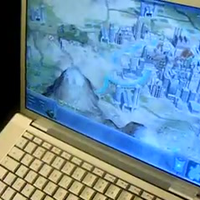}
\includegraphics[width=1cm]{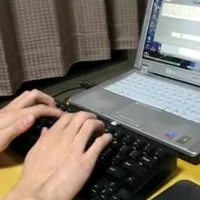}
\includegraphics[width=1cm]{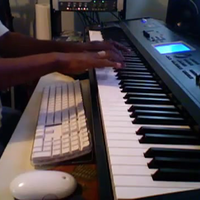}
& 
\includegraphics[width=1cm]{png/po2tcrG6KzM_2_8_0004.png}
\includegraphics[width=1cm]{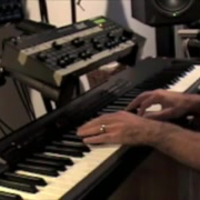}
\includegraphics[width=1cm]{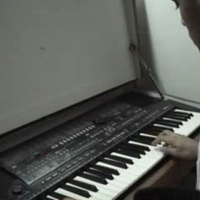}
\\


(2) Kids are playing in a pool.&&\\
\includegraphics[width=1cm]{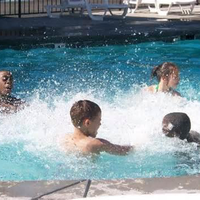}
\includegraphics[width=1cm]{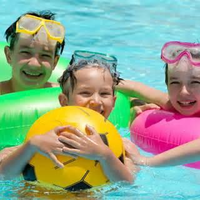}
\includegraphics[width=1cm]{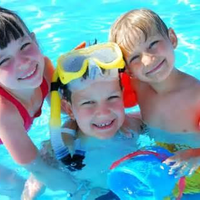}
\includegraphics[width=1cm]{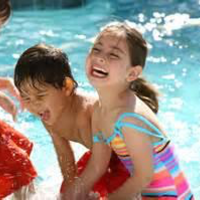}
\includegraphics[width=1cm]{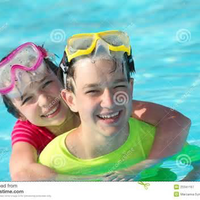}
&
\includegraphics[width=1cm]{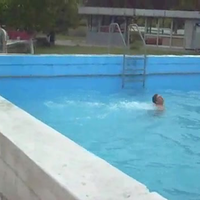}
\includegraphics[width=1cm]{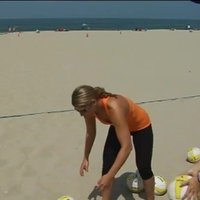}
\includegraphics[width=1cm]{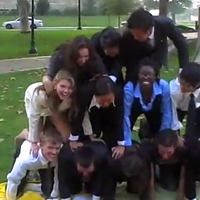}
& 
\includegraphics[width=1cm]{png/zS50h-a8RTg_3_9_0004.png}
\includegraphics[width=1cm]{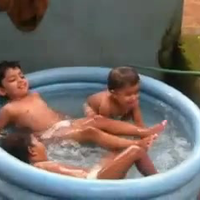}
\includegraphics[width=1cm]{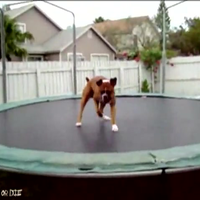}
\\

(3) A man is trimming fat from a roast.&&\\
\includegraphics[width=1cm]{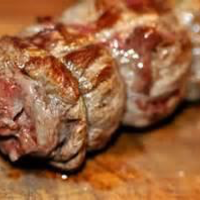}
\includegraphics[width=1cm]{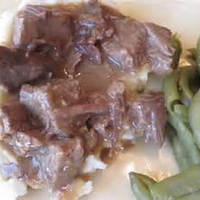}
\includegraphics[width=1cm]{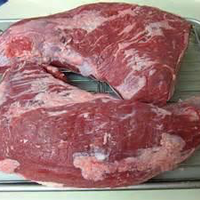}
\includegraphics[width=1cm]{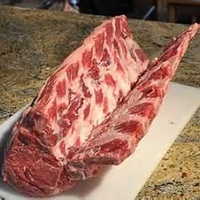}
\includegraphics[width=1cm]{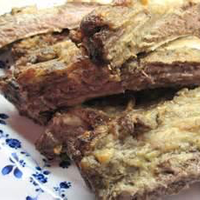}
&
\includegraphics[width=1cm]{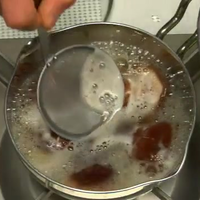}
\includegraphics[width=1cm]{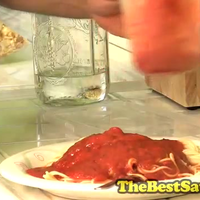}
\includegraphics[width=1cm]{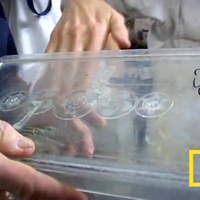}
& 
\includegraphics[width=1cm]{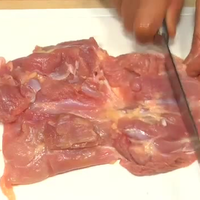}
\includegraphics[width=1cm]{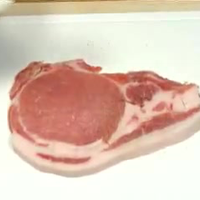}
\includegraphics[width=1cm]{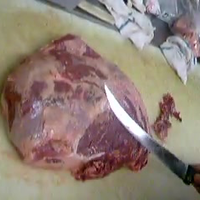}
\\
\multicolumn{1}{c}{} & \multicolumn{1}{c}{} & \multicolumn{1}{c}{}\\
\multicolumn{1}{c}{\textbf{Query}} & \multicolumn{1}{c}{\textbf{GoogLeNet+VI}} & \multicolumn{1}{c}{\textbf{GoogLeNet+ALL$_2$}}\\

(4) A boy is singing into a microphone.&&\\
\includegraphics[width=1cm]{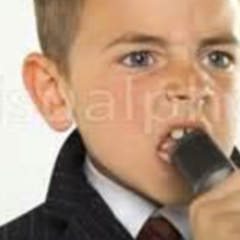}
\includegraphics[width=1cm]{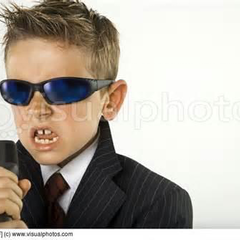}
\includegraphics[width=1cm]{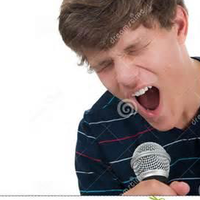}
\includegraphics[width=1cm]{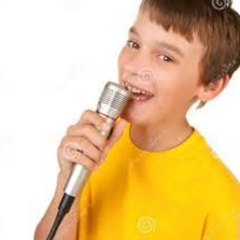}
\includegraphics[width=1cm]{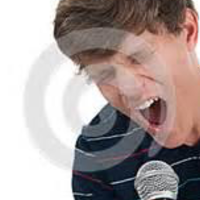}
&
\includegraphics[width=1cm]{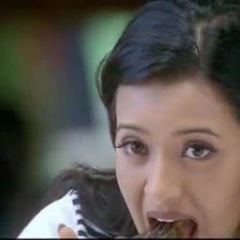}
\includegraphics[width=1cm]{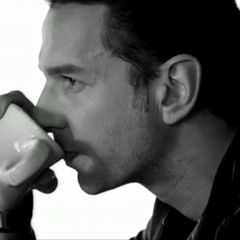}
\includegraphics[width=1cm]{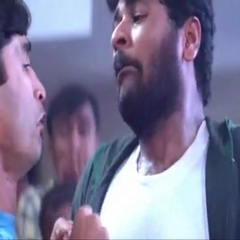}
& 
\includegraphics[width=1cm]{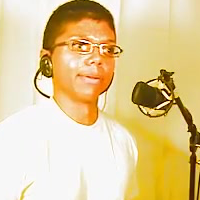}
\includegraphics[width=1cm]{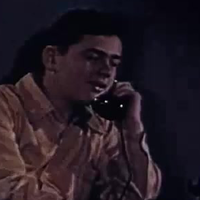}
\includegraphics[width=1cm]{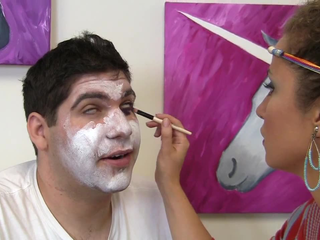}
\\

(5) A man shoots a shotgun.&&\\
\includegraphics[width=1cm]{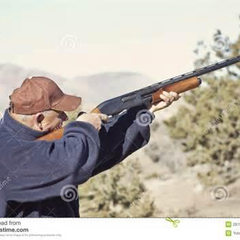}
\includegraphics[width=1cm]{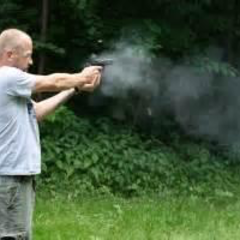}
\includegraphics[width=1cm]{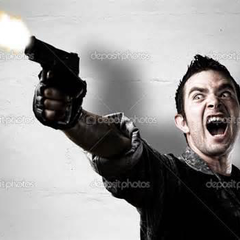}
\includegraphics[width=1cm]{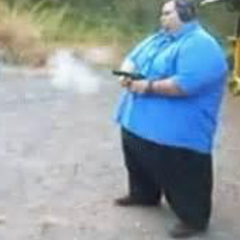}
\includegraphics[width=1cm]{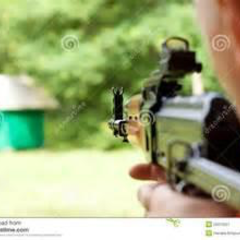}
&
\includegraphics[width=1cm]{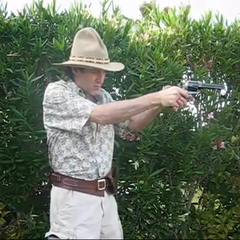}
\includegraphics[width=1cm]{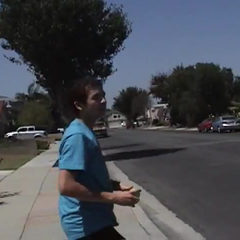}
\includegraphics[width=1cm]{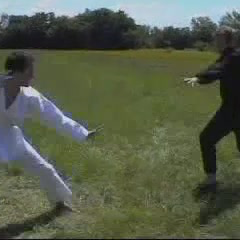}
& 
\includegraphics[width=1cm]{png/IAvBB2lv8iw_142_148_0004.png}
\includegraphics[width=1cm]{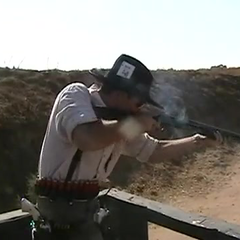}
\includegraphics[width=1cm]{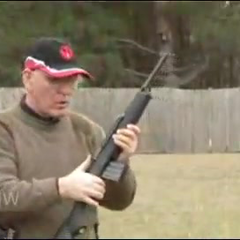}
\\

(6) A cat is pawing in a water bowl.&&\\
\includegraphics[width=1cm]{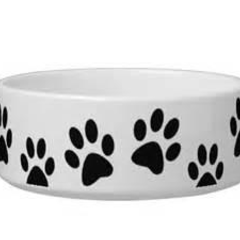}
\includegraphics[width=1cm]{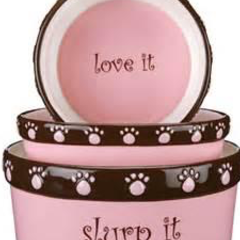}
\includegraphics[width=1cm]{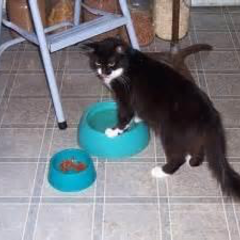}
\includegraphics[width=1cm]{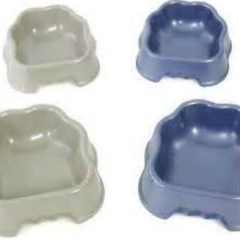}
\includegraphics[width=1cm]{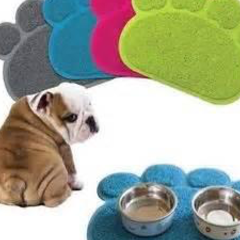}
&
\includegraphics[width=1cm]{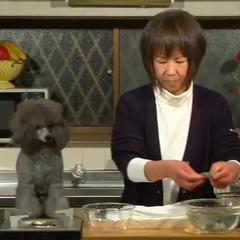}
\includegraphics[width=1cm]{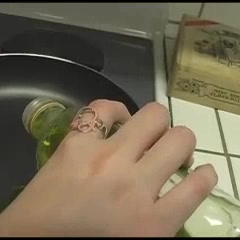}
\includegraphics[width=1cm]{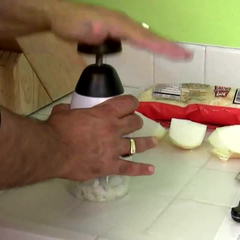}
& 
\includegraphics[width=1cm]{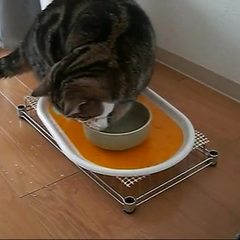}
\includegraphics[width=1cm]{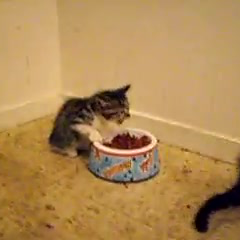}
\includegraphics[width=1cm]{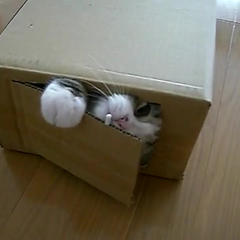}
\\
\end{tabular}
\caption{Examples of video retrieval results. Left: Query sentence and web images.  Center: Top-3 retrieved videos by GoogLeNet+VS and VI. Right: Top-3 retrieved videos by GoogLeNet+ALL$_2$. \label{fig:vid_ret}}
\end{figure}

Some examples of retrieved videos by GoogLeNet+VS, GoogLeNet+VI, and GoogLeNet+ALL$_2$ are shown in Fig. \ref{fig:vid_ret}.
These results suggest that web images reduced the ambiguity of queries' semantics by providing hints on their visual concepts.
For example, with sentence (1) ``A man is playing a keyboard,'' retrieval results of GoogleNet+VS includes two videos of a keyboard on a laptop as well as one on a musical instrument.
On the other hand, all top-3 results by GoogleNet+ALL$_2$ are about musical instruments.
Compared to GoogLeNet+VI, our full model obtained more videos with relevant content.
Moreover, the result of query (6) indicates that our model can recover from irrelevant image search results by combining a query sentence.

\begin{figure}[t!]
\centering
\includegraphics[width=0.95\linewidth]{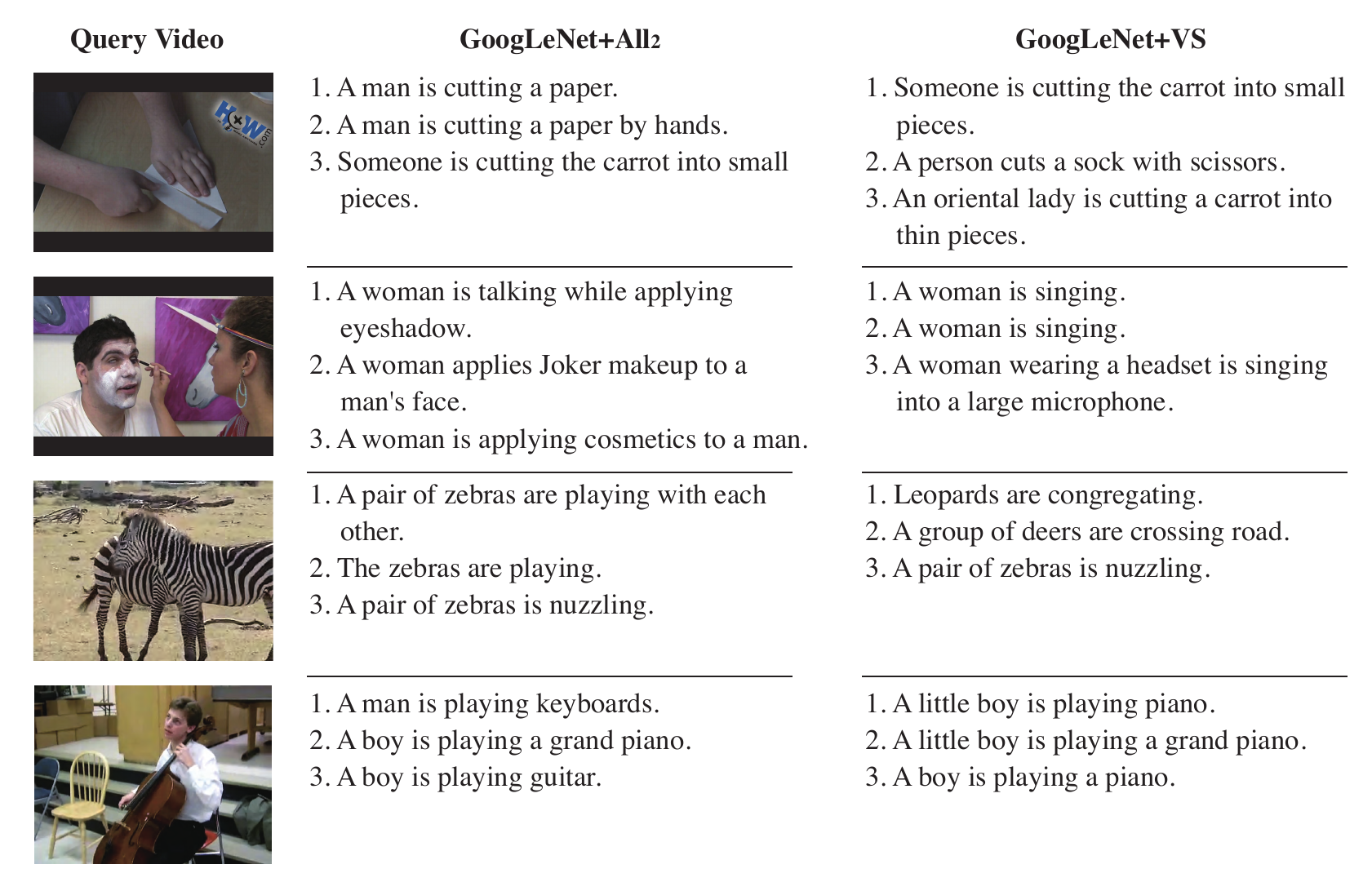}
\caption{Examples of top-3 retrieved sentences. Left: Query videos. Center: Top-3 retrieved sentences by GoogLeNet+ALL$_2$. Right: Top-3 retrieved sentences by GoogLeNet+VS.\label{fig:desc_ret}}
\end{figure}

Some examples of sentence retrieval results are shown in Fig.~\ref{fig:desc_ret}.
While our full model may retrieve sentences that disagree with query videos in details, most of the retrieved sentences are relevant to query videos.

\subsection{Comparison to Prior Work}
The approach for image and sentence retrieval by Kiros \etal~\cite{Kiros2015} applies linear transformations to CNN-based image and RNN-based sentence representations to embed them into a common space.
Note that their model was designed for the image and sentence retrieval tasks; thus, we extracted the middle frame as a keyframe and trained the model with pairs of a keyframe and a sentence.
Xu \etal~\cite{Xu2015} introduced neural network-based embedding models for videos and sentences.
Their approach embeds videos and SVO triplets extracted from sentences into an embedding space.
Kiros \etal's and Xu \etal's approaches are denoted by ST and DVCT, respectively.

Scores in Table \ref{tbl:ranking_score1} indicates that our model clearly outperformed prior work in both video and sentence retrieval tasks.
There is a significant difference in performance of DVCT and others.
ST and ours encode all words in a sentence, while DVCT only encodes its SVO triplets.
This suggests that using all words in a sentence together with an RNN is necessary to get good embeddings.

\section{Video Description Generation}
\begin{figure}[t!]
\begin{center}
\includegraphics[clip, width=0.75\linewidth]{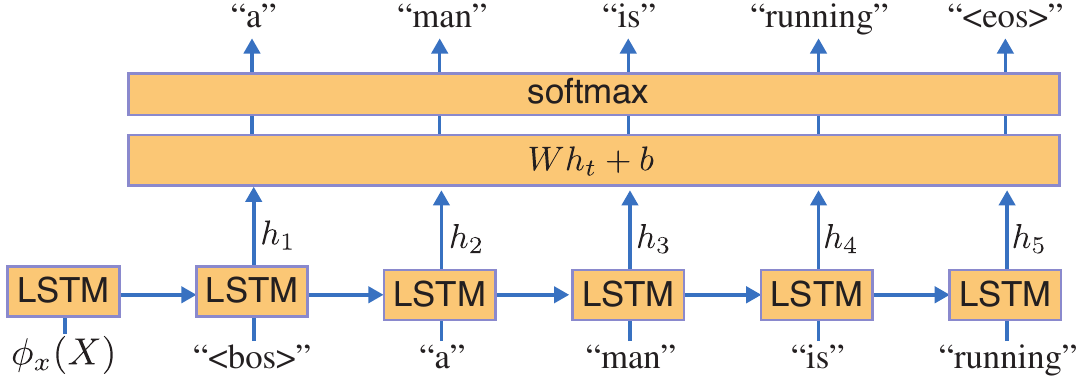}
\end{center}
 \caption{Illustration of the decoder model. ``$<$bos$>$'' is a tag denoting the beginning of a sentence, and ``$<$eos$>$'' is the end of a sentence.\label{fig:decoder_illust}}
\end{figure}
\begin{figure}[t]
\scriptsize
\centering
\begin{tabular}{p{.23\linewidth}@{\hspace{.45cm}}p{.23\linewidth}@{\hspace{.45cm}}p{.23\linewidth}}
\includegraphics[width=\linewidth]{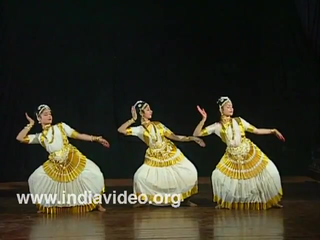}&
\includegraphics[width=\linewidth]{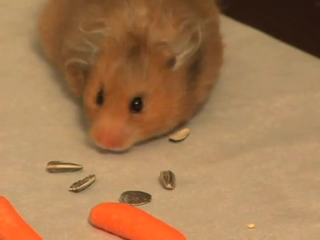}&
\includegraphics[width=\linewidth]{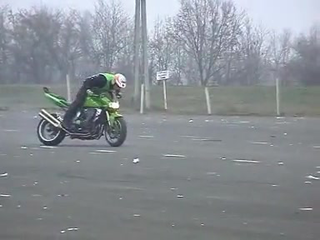}\\
Women are dancing.&
A hamster eats seeds.&
A man is riding a motorcycle.\\

\includegraphics[width=\linewidth]{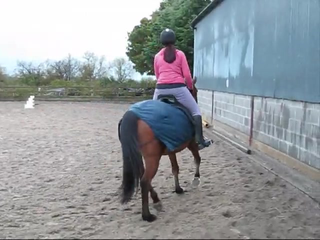}&
\includegraphics[width=\linewidth]{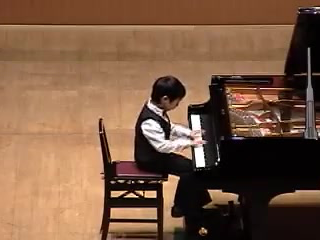}&
\includegraphics[width=\linewidth]{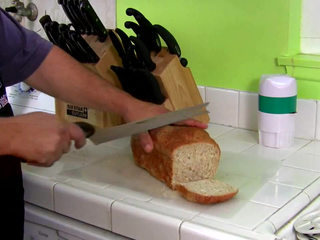}\\
A woman is riding a horse.&
A man is playing a piano.&
\tc{A man is slicing a potato.}\\


\end{tabular}
\caption{Sentences generated from our video embeddings. The sentence in red is a failure.\label{fig:desc_gen}}
\end{figure}

\begin{table}[t!]
\caption{Scores of generated sentences. TVNL+Extra Data is the TVNL model pre-trained on the Flickr30k \cite{TACL229} and the COCO2014 \cite{capeval2015} datasets.\label{tbl:gen_score}}
\begin{center}
\begin{tabular}{r|C{2cm}C{2cm}C{2cm}}
\hline
\multicolumn{1}{C{2cm}|}{\textbf{Models}} & \multicolumn{1}{C{2cm}}{\textbf{CIDEr}} & \multicolumn{1}{C{2cm}}{\textbf{BLEU}} & \multicolumn{1}{C{2cm}}{\textbf{METEOR}} \\ \hline
TVNL \cite{Venugopalan2014}                                 &                   -                 & 31.19                                & 26.87                               \\
TVNL+Extra Data     &                  -                  & 33.29                                & 29.07                               \\
DVETS \cite{Yao2015}                                & 51.67                               & 41.92                                & 29.60                               \\
Ours                                 & 41.62                               & 33.69                                & 28.47                               \\ \hline
\end{tabular}
\end{center}
\end{table}

Automatic description generation for images \cite{Vinyals2015,Fang2015} and videos \cite{Rohrbach2013,Venugopalan2014,Yao2015} is another task to associate images or videos with sentences. 
As an application of our models, we performed the description generation task using our video embeddings.
To analyze the information encoded by our video embedding, we trained a decoder that produces descriptions from our video embeddings.
A basic approach for description generation is to use long-short term memory (LSTM) that produces a sequence of probabilities over a vocabulary conditioned on visual representations \cite{Vinyals2015,Venugopalan2014}. 
We trained an LSTM as a decoder of video embeddings (Fig. \ref{fig:decoder_illust}).
The decoder predicts the next word based on word vector $w_{t}$ at each time step $t$ as:
\begin{eqnarray}
[a_t\ i_t\ f_t\ o_t]^T &=& W_{\mathrm{u}} w_{{t}} + b_{\mathrm{u}} + W_{\mathrm{l}} h_{t-1},\\
c_t &=& \tanh(a_t)\sigma(i_t)+c_{t-1}\sigma(f_t),\\
h_t &=& \tanh(c_t)\sigma(o_t),\\
p_t &=& \mathrm{softmax}( W_{\mathrm{p}}h_t+b_{\mathrm{p}})
\end{eqnarray}
where $W_\mathrm{u}, W_\mathrm{l} \in \mathrm{R^{4d_w\times d_w}}$ and $b_u \in \mathrm{R}^{4d_\mathrm{w}}$ are parameters of the LSTM, and $[a_t\ i_t\ f_t\ o_t]^T$ is a column vector that is a concatenation of $a_t, i_t, f_t, o_t \in \mathrm{R}^{d_\mathrm{w}}$.
The matrix $W_\mathrm{p}$ and the vector $b_\mathrm{p}$ encode the hidden state into a vector with the vocabulary size.
The output $p_t$ is the probabilities over the vocabulary.
We built a vocabulary consisting of all words in the YouTube dataset and special tags, \ie, begin-of-sentence (``$<$bos $>$'') and end-of-sentence (``$<$eos $>$'').
The generative process is terminated when ``$<$eos $>$'' is produced.
We trained the decoder using the YouTube dataset.
We computed the video embedding $\phi_{\mathrm{v}}(X)$ using GoogLeNet+ALL$_2$ as an input to the LSTM at $t=0$.
We trained the decoder by minimizing the cross entropy loss.
During training, we fixed the parameters of our embedding models.

Figure \ref{fig:desc_gen} shows generated sentences.
Although video embeddings were trained for retrieval tasks and not finetuned for the decoder, we observed that most generated sentences were semantically relevant to their original videos.

We evaluated generated sentences with the COCO description evaluation.
We found that the scores were comparable to prior work (Table \ref{tbl:gen_score}).
This indicates that our model efficiently encoded videos, maintaining their semantics.
Moreover, this result suggests that our embeddings can be applied to other tasks that require joint representations of videos and sentences.

\section{Conclusion}
We presented a video and sentence retrieval framework that incorporates web images to bridge between sentences and videos.
Specifically, we collected web image search results in order to disambiguate semantics of a sentence.
We developed neural network-based embedding models for video, sentence, and image inputs which fuses sentence and image representations.
We jointly trained video and sentence embeddings using the YouTube dataset.
Our experiments demonstrated the advantage of incorporating additional web images, and our approach clearly outperformed prior work in the both video and sentence retrieval tasks.
Furthermore, by decoding descriptions from video embeddings, we demonstrated that rich semantics of videos were efficiently encoded in our video embeddings.
Our future work includes developing a video embedding that considers temporal structures of videos.
It would be also interesting to investigate what kind of sentences benefit from image search results, and how to collect efficient images.
\clearpage

\bibliographystyle{splncs03}
\bibliography{egbib}
\end{document}